\documentclass[a4paper,twoside]{article}

\usepackage{epsfig}
\usepackage{subcaption}
\usepackage{calc}
\usepackage{amssymb}
\usepackage{amstext}
\usepackage{amsmath}
\usepackage{amsthm}
\usepackage{xcolor}
\usepackage{multicol}
\usepackage{pslatex}
\usepackage{apalike}
\usepackage[bottom]{footmisc}
\usepackage{SCITEPRESS}     % Please add other packages that you may need BEFORE the SCITEPRESS.sty package.
\usepackage[para,online,flushleft]{threeparttable}

\begin{document}

\title{Characterizing Speed Performance of\\ Multi-Agent Reinforcement Learning \thanks{This work is supported by the U.S. National Science Foundation (NSF) under grant CNS-2009057 and the Army Research Lab (ARL) under grant W911NF2220159.}}

\author{\authorname{Samuel Wiggins\sup{1}\orcidAuthor{0000-0002-0069-8213}, Yuan Meng\sup{1}\orcidAuthor{0000-0001-6468-8623}, Rajgopal Kannan\sup{2}\orcidAuthor{0000-0001-8736-3012} and Viktor Prasanna\sup{1}\orcidAuthor{0000-0002-1609-8589}}
\affiliation{\sup{1}Ming Hsieh Department of Electrical and Computer Engineering, University of Southern California, Los Angeles, USA}
\affiliation{\sup{2}DEVCOM Army Research Lab, Los Angeles, USA}
\email{\{wigginss, ymeng643, prasanna\}@usc.edu, rajgopal.kannan.civ@army.mil}
}
\abstract{Multi-Agent Reinforcement Learning (MARL) has achieved significant success in large-scale AI systems and big-data applications such as smart grids, surveillance, etc. Existing advancements in MARL algorithms focus on improving the rewards obtained by introducing various mechanisms for inter-agent cooperation. However, these optimizations are usually compute- and memory-intensive, thus leading to suboptimal speed performance in end-to-end training time. In this work, we analyze the speed performance (i.e., latency-bounded throughput) as the key metric in MARL implementations. Specifically, we first introduce a taxonomy of MARL algorithms from an acceleration perspective categorized by (1) training scheme and (2) communication method. Using our taxonomy, we identify three state-of-the-art MARL algorithms - Multi-Agent Deep Deterministic Policy Gradient (MADDPG), Target-oriented Multi-agent Communication and Cooperation (ToM2C), and Networked Multi-agent RL (NeurComm) - as target benchmark algorithms, and provide a systematic analysis of their performance bottlenecks on a homogeneous multi-core CPU platform. We justify the need for MARL latency-bounded throughput to be a key performance metric in future literature while also addressing opportunities for parallelization and acceleration.
}
\onecolumn \maketitle \normalsize \setcounter{footnote}{0} \vfill

\section{\uppercase{Introduction}} 
\label{sec:introduction}  
Reinforcement Learning (RL) is a crucial technique for model development and algorithmic innovation in data science. Multi-agent Reinforcement Learning (MARL), an extension of single-agent RL, has shown advancement in many big-data application domains such as e-commerce \cite{choi2022marl} and surveillance in large-scale systems \cite{ivan2020methods}. Compared to single-agent RL, MARL introduces agent-to-agent interactions using algorithm-specific communication protocols. MARL also poses challenges in data processing since each agent generates a large amount of data, which is high-dimensional and correlated with the data from other agents. The speed performance of processing data in MARL is critical as MARL training is extremely time-consuming. In Deep MARL, using Deep Neural Network (DNN) as a policy model, multiple autonomous agents interact with a shared environment to achieve a joint goal. The majority of state-of-the-art MARL algorithms utilize DNNs, so we use MARL and Deep MARL interchangeably in this paper.

\begin{figure}[ht]
    \centering
    \includegraphics[width=7.5cm]{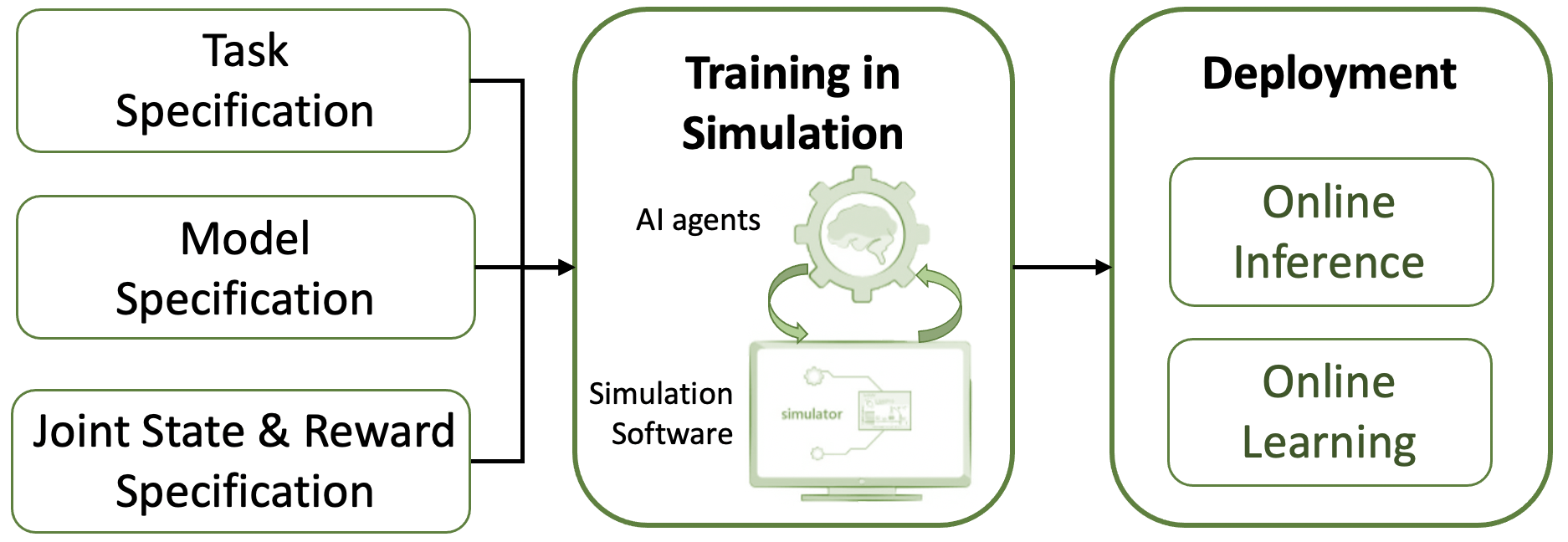}
    \caption{MARL application workflow}
    \label{fig:Workflow}
\end{figure}

Figure \ref{fig:Workflow} shows a high-level diagram of Deep MARL in a real-world scenario. A typical workflow consists of two major stages: Training in Simulation and Deployment. In this paper, we focus on Training using a simulation, which is the critical process before deploying a MARL system into its actual physical environment. This Training in Simulation process is the most time-consuming part of MARL system development, often taking days to months to train an acceptable model \cite{rl_survey}. There has been extensive work accelerating this stage for single-agent RL. However, Training in Simulation for MARL systems brings non-trivial computational challenges compared to single-agent scenarios due to the requirement of facilitating inter-agent communications. There is a lack of literature on parallelization and acceleration in a multi-agent setting stemming from the omission of MARL system execution time as a performance metric. In addition to focusing on maximizing cumulative reward through novel algorithmic optimizations,
we argue that optimizing the MARL system execution time through parallelization and acceleration should be considered.

In practice, the efficiency of developing and deploying high-speed MARL systems is dependent upon several factors such as the dependency requirement within RL execution loops, the various Computation intensities of MARL primitives, and the suitability of the memory hierarchy. However, widely-used homogeneous platforms (i.e., CPUs) cannot satisfy all of the above factors simultaneously, leading to various time overheads that prevent the MARL system from achieving theoretical peak throughput.

In this paper, we provide a systematic empirical analysis of the performance of three state-of-the-art MARL algorithms - MADDPG, ToM2C, and NeurComm on CPUs. 
To the best of our knowledge, this is the first work that analyzes MARL algorithms from a speed performance and acceleration point of view.
The main contributions of the paper are:
\begin{itemize}
    \item We provide a taxonomy of MARL algorithms from an acceleration point of view, summarize their parallelization parameters and highlight the computation characteristics of each category. 
    \item We compare the timing breakdown and performance scalability with respect to the parallelization parameters of state-of-the-art MARL implementations on a multi-core CPU platform.
    \item We provide reasoning into why MARL system execution time should be considered a key performance metric, and show new acceleration challenges and opportunities that emerge.
\end{itemize}

\vspace{-10pt}
\section{\uppercase{Background}}
\vspace{-10pt}
\subsection{Multi-Agent Reinforcement Learning}

We formulate the decision-making problem in MARL as an $n-$agent Markov game \cite{shapley1953stochastic}, which can be defined as a tuple $\left(\mathcal{S}, \mathcal{A}^1, \ldots, \mathcal{A}^n, R^1, \ldots, R^n, \mathcal{T}, \gamma\right)$. In this tuple, ${1...n}$ is the set of agents, $\mathcal{S}$ is the state space, $\mathcal{A}^i$ is the action space of agent $i$, $R^i: \mathcal{S} \times \mathcal{A} \mapsto \mathbb{R}$ is the reward function
of agent $i$, $\mathcal{T}: \mathcal{S} \times \mathcal{A} \mapsto \Delta(\mathcal{S})$ denotes the transition probability of each state-action pair to another state, $\gamma \in[0,1)$ is a discount factor for future rewards. For agent $i$, we denote its policy as a probability distribution over its action space $\pi^i: \mathcal{S} \rightarrow \Delta\left(\mathcal{A}^i\right)$, where $\pi^i\left(a_t \mid s_t\right)$ is the probability of taking action $a_t$ upon state $s_t$ at a certain time step $t$. By denoting the other agents' actions as $a_t^{-i}=\left\{a_t^j\right\}_{j \neq i}$, we formulate the other agents' joint policy as $\pi^{-i}\left(a_t^{-i} \mid s_t\right)=\Pi_{j \in\{-i\}} \pi^j\left(a_t^j \mid s_t\right)$. At each
time step, actions are taken simultaneously. Each agent $i$ aims
at finding its optimal policy to maximize the expected return
(cumulative reward), defined as
\begin{multline}
\label{eq:obj}
    \mathbb{E}_{\left(s_t, a_t^i, a_t^{-i}\right) \sim \mathcal{T}, \pi^i, \pi^{-i}}\left[\sum_{t=1}^{\infty} \gamma^t R^i\left(s_t, a_t^i, a_t^{-i}\right)\right]
\end{multline}
From Equation \ref{eq:obj}, the optimal policy of agent $i$ depends not only on its own policy but also on the behaviors of other agents. Depending on the assumptions of how agents communicate, MARL policy optimization can be categorized into several scenarios which will be further discussed in Section \ref{sec:taxonomy}.

\begin{figure}[h]
    \centering
    \includegraphics[width=7.5cm]{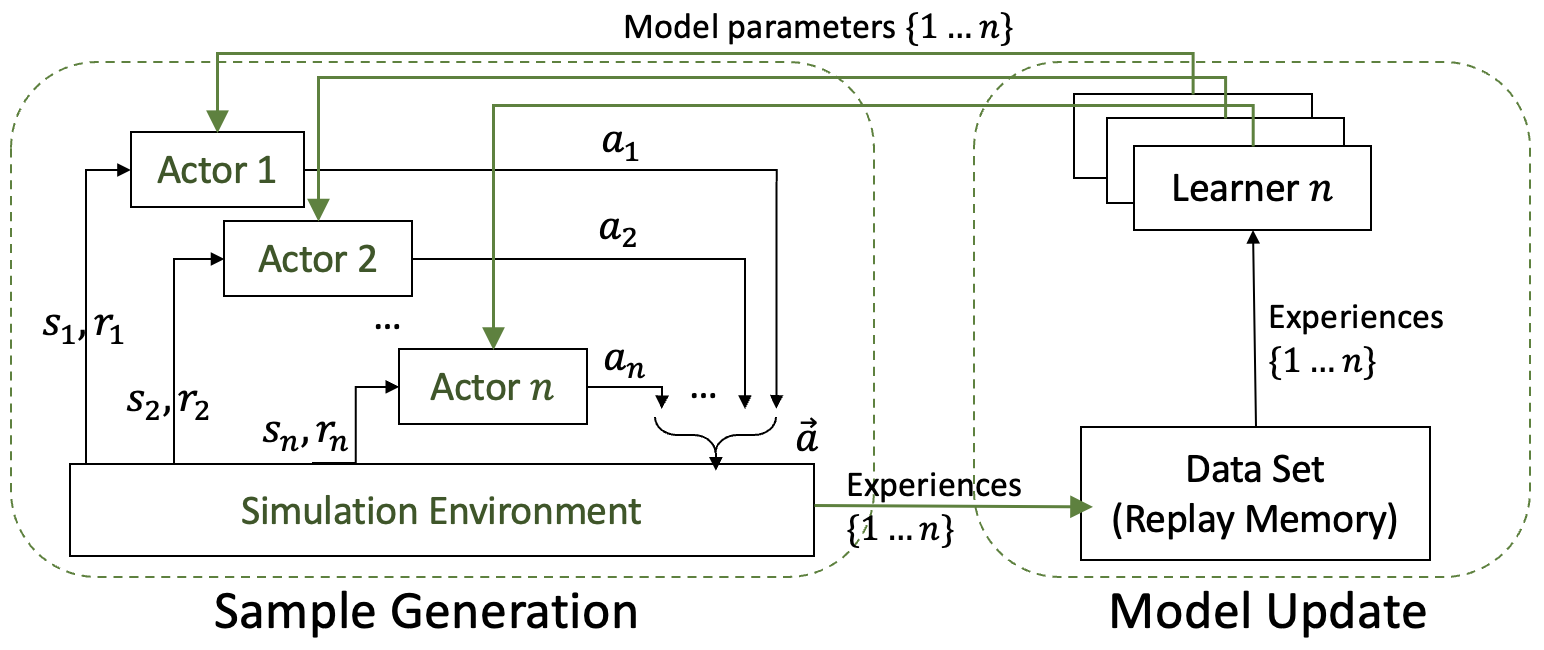}
    \caption{Training in Simulation}
    \label{fig:TinS}
\end{figure}

As shown in Figure \ref{fig:TinS}, each iteration of the Training-in-Simulation process can be divided into the Sample Generation and Model Update phases.
 In the Sample Generation phase, the actor of each agent takes the environment state as input to the DNN policy and inference an action. The joint action of all the agents is simultaneously executed to transit to a new state. This process is repeated until a preset maximum trajectory length value is achieved or a terminal state is reached. At the end of the Sample Generation phase, the agents produce a batch of experiences $\{s_t, \overrightarrow{a_t}, r_t, s_{t+1}\}$. Then, in the Model Update phase, the learner of each agent samples (mini-)batches of experiences to train the DNN policy and propagate the updated policy back to the actor of the same agent. 
\vspace{-5pt}
\subsection{Related Work}
\vspace{-10pt}
\subsubsection{Reinforcement Learning Acceleration}
There is a plural of work for parallelization and acceleration of single-agent RL. On general-purpose platforms (i.e., CPU and GPU), existing work adopts coarse-grained parallelism by deploying and distributing multiple actors and learners for the same agent \cite{zhang2021parallel,liang2018rllib}, as well as fine-grained parallelism within each actor or learner for simulations, batched inference, and training \cite{rocki2011parallel}. Additionally, specialized hardware accelerators targeting specific single-agent RL algorithms are developed \cite{cho2019fa3c,meng2021ppoaccel,hpec}. However, not all of these acceleration techniques can be trivially adapted to MARL. This is because MARL poses different challenges. For example, the latency bottleneck of MARL may become the overhead from coordination between agents, which is nonexistent in single-agent RL. 

\subsubsection{MARL Implementation: Summary and Challenges}
 Compared to single-agent RL, MARL leads to additional problems in terms of increased state-action complexity, partial observability from each agent, and coordination requirement in cooperative settings. Many MARL algorithms have been proposed and implemented to address these problems, where their major contributions are targeted toward learning to communicate effectively between RL agents. 
 \cite{lowe2017multi,jiang2018graph} allow agents to perform in the environment in a decentralized way with copies of a shared neural network and enable instantaneous communication between all the agents for training the policies.
\cite{chu2020multi,wang2021tom2c} propose dedicated compute modules to learn using a shared graph for agents to decide whether and with whom to communicate.
These works focus on increasing the sample efficiency and communication efficiency to improve the convergence rate to a high joint reward.
They are implemented on multi-core CPUs, and computationally-intensive workloads such as neural network training are offloaded to a GPU. However, these implementations are usually highly time-consuming. A unique challenge that impedes the speed performance of MARL is the \textbf{overheads from communication}.
The compute primitives and memory modules for learning communication, including message generation (e.g., Recurrent neural networks \cite{IS}) and message propagation (e.g., Sparse linear algebra based on Graph Neural Networks \cite{jiang2018graph}), are tightly coupled with agent policy training even though they have highly dissimilar compute patterns and memory operations compared to policy training. This leads to various computation overheads that lower the effective hardware utilization on CPU, which in turn impedes the policy execution and training speed. In this paper, we aim to characterize these overheads and identify acceleration opportunities in MARL training.
\section{\uppercase{A Taxonomy of MARL Algorithms}}
\label{sec:taxonomy}
\begin{figure}[h]
    \centering
    \includegraphics[width=7.5cm]{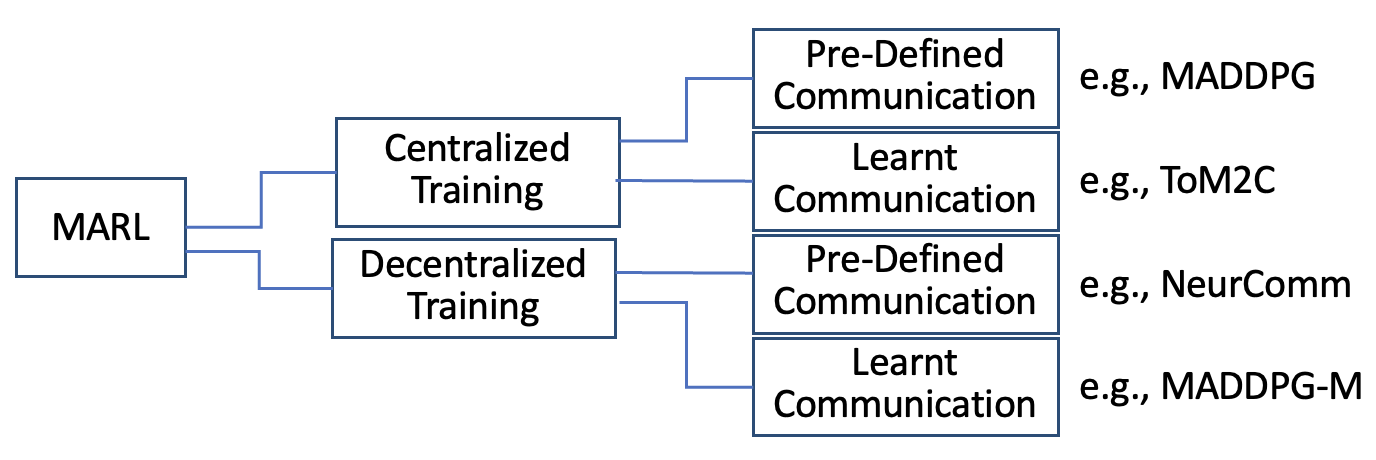}
    \caption{MARL Taxonomy}
    \label{fig:taxonomy}
\end{figure}

We illustrate a structural way of categorizing MARL algorithms based on their computation characteristics and major acceleration requirements. This is shown in two dimensions: training scheme (centralized vs. decentralized) and communication method (pre-defined vs. online learnt). Figure \ref{fig:taxonomy} shows example algorithms in each category. 

\vspace{-5pt}
\subsection{Training Scheme}
\vspace{-5pt}
Based on the requirement of MARL applications, each agent can learn in a decentralized way utilizing respective local experiences, or in a centralized manner where all the agents share a global pool of model parameters and joint experiences.

\textbf{Centralized Training - Decentralized Execution (CTDE):} In the Model Update phase under Centralized Training - Decentralized Execution (CTDE), a centralized controller or learner is responsible for coordinating the policy training of all the agents. The centralized controller has access to global observations and rewards. Once the training is complete, however, each agent executes its actions independently, based on its own local observations, without any further communication with the centralized controller. Communication is required during Model Update and is usually implemented using a shared memory architecture on modern CPUs and GPUs \cite{wang2021tom2c,lowe2017multi}. 

\textbf{Decentralized Training:} In Decentralized Training, each agent in the system learns its own policy without direct access to the experiences of other agents in the system. From the perspective of a single agent, the environment becomes non-stationary due to the co-adaption of the other agents, such that the policy learned can have a mismatched expectation about the other agents' policies. Without a centralized controller, the problems of partial observability and non-stationarity become more complex. Therefore, communication that allows agents to exchange information becomes a critical factor in the coordination of their behaviors and overcoming the partial observability and non-stationarity problems \cite{wang2022model}. 

\subsection{Communication Method}
Communication Policy defines how to make the decisions of `whom' and `when' to communicate with and enable message transferring across agents. A communication policy can be either pre-defined or learnt. 

\textbf{Pre-Defined Communication:}
Pre-defined communication specifies a fixed communication protocol and message format before the training process begins. Agents may perform all-to-all communication \cite{lowe2017multi}, or capture the communication paths using a pre-defined graph that is associated with the given application or environment \cite{chu2020multi,jiang2018graph}.

\textbf{Learnt Communication:}
Learnt communication offers generalization abilities to more scenarios and has become increasingly popular due to its flexibility. One popular method is to adaptively learn the evolving node and edge features of the multi-agent system represented as a dynamic graph \cite{wang2021tom2c}.
\section{\uppercase{Latency-Bounded Throughput of MARL Algorithms: An Analysis}}
\subsection{Metrics and Parallel Parameters}
\subsubsection{Latency-Bounded Throughput}
Given a particular algorithm and benchmark that assumes a fixed sample complexity (i.e., the number of iterations needed to reach a certain reward), the primary metric for measuring MARL training speed is the throughput in terms of the number of Iterations executed Per Second ($IPS$). As the iterations in Training-in-Simulation are sequential by nature, it is hard to overlap consecutive iterations for faster convergence \cite{cho2019fa3c,meng2021ppoaccel}, so minimizing the latency of each iteration is critical for high $IPS$ throughput. Based on the Sample Generation and Model Update phases in each iteration (Figure \ref{fig:TinS}), we formulate the throughput $IPS$ as:
\begin{equation}
IPS =\frac{1}{T_{\text {iteration }}}=\frac{1}{\circledast\left(T_{S G}, T_{\text {MU}}\right)} \text {, }
\end{equation}
where $T_{S G}$ and $T_{\text {MU}}$ are the execution times of Sample Generation and Model Update phases, respectively. The operator $\circledast(x,y)$ can be $x+y$ or $max(x,y)$ based on whether the algorithm is on-policy (where Model Update is strictly dependent on Sample Generation in each iteration) or off-policy (where Sample Generation and Model Update phases can be overlapped). 

In the following subsections, we provide a detailed analysis of the main computation primitive latency and throughput using representative algorithms in different categories of MARL. We will show the time breakdown and performance scalability with respect to a few key parallel parameters.
\subsubsection{Key Parallel Parameters}

A MARL algorithm is defined using a plural of hyper-parameters, where a subset of these hyper-parameters affects the parallelism, data processing speed, and latency-bounded throughput of the system.
We define the following terms relevant to the key parallel parameters that are common in all categories of MARL:
\begin{itemize}
    \item \textbf{Agent(s):} Agents execute by interacting with the environment and improving their policies.
    \item \textbf{Actor(s):} An actor is a process used by an agent to perform the Sample Generation phase. An actor infers on the agent's policy to perform an action in a simulation environment. Each agent can spawn multiple actors implemented on multiple rollout thread(s) defined below.
    \item \textbf{Rollout thread(s):} 
    A rollout thread is a CPU thread used to execute the actor process. 
    A rollout thread has a local copy of the environment and executes independently of the other actor/rollout thread(s). Typically, each agent deploys multiple actors, and each actor is mapped to a rollout thread. 
    \item \textbf{Learner(s):} A learner is the process used by an agent to train its policy in the Model Update phase. 
    In CTDE, the learner is typically shared by all the agents, so the number of learners in the MARL system is 1. In Decentralized training, the number of learners is equal to the number of agents.
\end{itemize}

\subsection{Centralized Training with Knowledge Concatenation}
\label{sec:maddpg}
Multi-Agent Deep Deterministic Policy Gradient (MADDPG) falls under the Centralized Training Decentralized Execution category with pre-defined all-to-all communication using knowledge concatenation \cite{lowe2017multi}. In MADDPG, each agent is composed of four DNN models: (1) The Policy network (i.e., decentralized actor) executes on the environment using only local information independent of other agents. (2) The Value network (i.e., centralized critic) helps train the actor's policy. Although it is considered centralized, each agent still trains its own value network. The training is centralized in the sense that the inputs to each value network depend on the action and observations (i.e., transition information) of all agents. (3,4) The Target Policy and Target Value network are used for training stability.

In the MADDPG implementation, multiple rollout threads are deployed to sample the environment using the policies from the decentralized actors, where the transition information is stored in a global replay buffer. Multiple training threads can then use this information to train the various DNN networks described above.

\begin{figure}[ht]
    \centering
    \includegraphics[width=7.5cm]{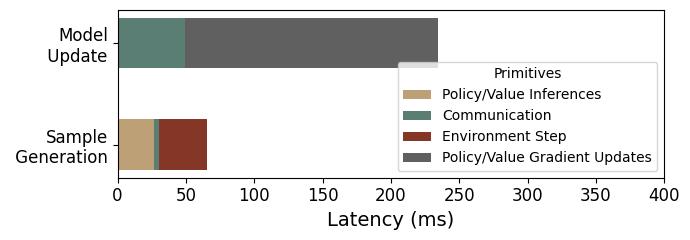}
    \caption{MADDPG Training in Simulation Breakdown (8 rollout threads, 1 training thread)}
    \label{fig:maddpg_brkd}
\end{figure}

During Sample Generation, each agent samples an action using its policy and executes it in the environment. The latency breakdown among Policy Inference, Communication, and Environmental Steps in one iteration is displayed in Figure \ref{fig:maddpg_brkd}. During Model Update, training occurs in an actor-critic fashion similar to the single-agent variant of MADDPG: Deep Deterministic Policy Gradient (DDPG) \cite{lillicrap2015ddpg}. Overall, we observe that communication in the sample generation and model update phase is not the dominating factor in terms of latency during Training in Simulation, which is attributed to the straightforward insertion and concatenation of transition information. 

\begin{figure}[ht]
    \centering
    \includegraphics[width=7.5cm]{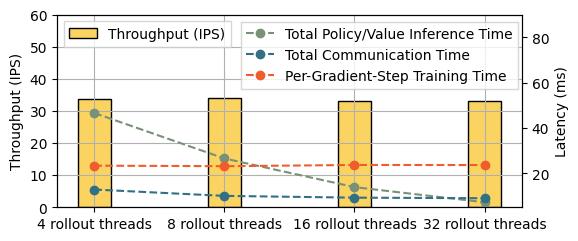}
    \caption{MADDPG Training Throughput Scalability varying rollout threads}
    \label{fig:maddpg_scale}
\end{figure}

We observe that the actor/critic policy and value updates are the dominating factor in terms of latency when varying the number of rollout threads as seen in Figure \ref{fig:maddpg_scale}. In the Sample Generation phase, the fixed amount of data generation takes a shorter amount of time since multiple parallel actors sample copies of the environment simultaneously while adding to the replay buffer. During the Model Update phase, latency for each gradient update step remains unchanged with respect to the number of rollout threads. This is expected given that training for each agent happens using a single thread sequentially regardless of the number of rollout threads. Increasing the number of rollout threads does increase the number of gradient steps needed to complete the model update phase in MADDPG. These additional steps cause the overall system throughput ($IPS$) to remain consistent. 

\subsection{Centralized Training with Learnt Communication}
\label{sec:tom2c}

ToM2C (Target-oriented Multi-agent Communication and Cooperation) is a recent work under the CTDE training scheme \cite{wang2021tom2c}.
In ToM2C, each agent learns its actions based on its local observation space along with inference of the mental states of others and learns the communication paths for enhancing cooperation.
Specifically, it comprises four functional DNN models: (1) The Observation Encoder is a fully-connected DNN layer that encodes the agent's local observation into a single feature using a weighted sum mechanism. (2) The Theory of Mind network (ToM Net) is a 2-layer Multi-Layer Perceptron on a Gate Recurrent Unit that estimates the joint intention of all the other agents. (3) The Message Sender is a Graph Neural Network (GNN) that uses the ToM Net output to decide the communication graph structure. (4) The Decision Maker (i.e., policy model) learns the actions once the agent receives all the messages based the communication graph returned by the Message Sender of all other agents.

In the ToM2C implementation, multiple actors on different rollout threads are used to generate datasets and store them in a shared buffer; a single training thread then uses the data from the shared buffer to update the policy, ToM Net, and GNN described above.

\begin{figure}[ht]
    \centering
    \includegraphics[width=7.5cm]{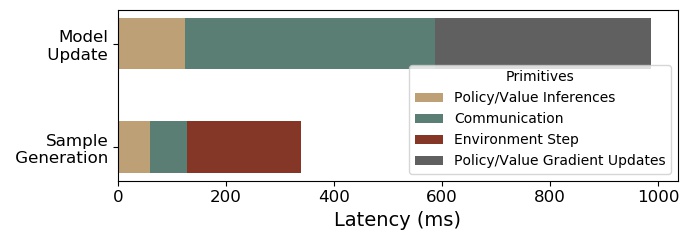}
    \caption{ToM2C Training in Simulation Breakdown (6 rollout threads, 1 training thread)}
    \label{fig:tom2c_brkd}
\end{figure}

During Sample Generation, each actor sequentially performs inference through the four DNN models. The time breakdown among Policy Inference, Communication, and Environmental Interaction in one iteration is shown in Figure \ref{fig:tom2c_brkd}. During Model Update, the training is performed in an actor-critic manner \cite{mnih2016asynchronous}, where a centralized critic (i.e., Value network) shared by all the agents obtain global feature of the states to compute their values and facilitate the training of the policy network (i.e., the Decision Maker). 

Overall, different from MADDPG which also follows the CTDE scheme, in ToM2C we observe a non-trivial overhead from learning of communication in the training pipeline. This is due to the added computation complexity from inferring other agents' intentions and deciding the communication paths using a graph representation of the global states of all the agents. On the other hand, the overheads from communication during agents execution (Sample Generation) are smaller compared to that in the Model Update, since the decentralized execution only requires each actor to perform one-pass inference on its own models.

\begin{figure}[ht]
    \centering
    \includegraphics[width=7.5cm]{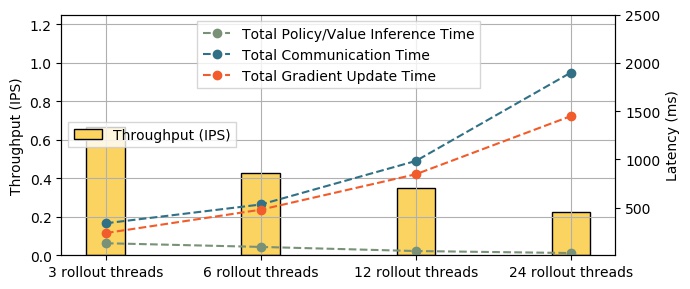}
    \caption{ToM2C Training Throughput Scalability varying the number of rollout threads}
    \label{fig:tom2c_scale_w}
\end{figure}
\begin{figure}[ht]
    \centering
    \includegraphics[width=7.5cm]{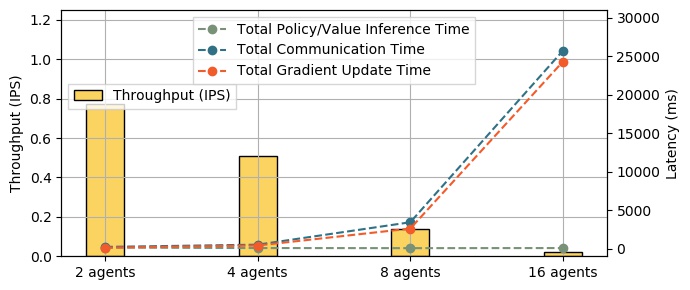}
    \caption{ToM2C Training Throughput Scalability varying the number of agents}
    \label{fig:tom2c_scale_a}
\end{figure}

In Sample Generation, adding rollout threads increases the data generation throughput (number of samples generated in unit time) without increasing the total latency needed for each actor to collect data, as shown in Figure \ref{fig:tom2c_scale_w}.
The actors and the learner use separate threads and only synchronize through a shared buffer; the Sample Generation and Model Update phases can run concurrently. Therefore, the latency of actors is completely hidden by the training process, such that increasing the number of rollout threads does not lead to higher system throughput ($IPS$).  

Figure \ref{fig:tom2c_scale_a} shows the execution time and $IPS$ scalability with increasing the number of agents cooperating in the same environment, with a fixed number of rollout threads serving each agent. We observe a faster rate of latency scaling (at approximately $8\times$ increase in gradient update time as the number of agents doubles). This factor is contributed by both the increased state size (processed state size doubles as the number of agents doubles) and the increased communication (worst-time communication complexity is the square of the number of agents). This means both algorithmic optimization and hardware acceleration for communication reduction are needed to increase the scalability of communication-learning MARL systems to a large number of agents.

\subsection{Decentralized Training with Pre-Defined Graph Communication}
\label{sec:neurcomm}

\begin{figure}[h]
    \centering
    \includegraphics[width=7.5cm]{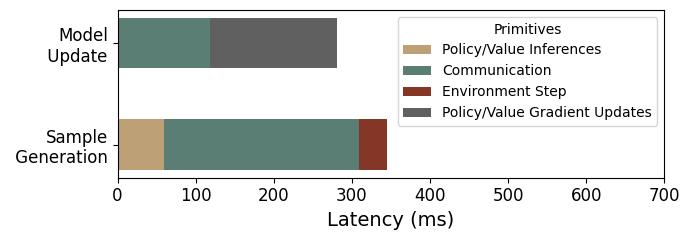}
    \caption{NeurComm Training in Simulation Breakdown (8 agents)}
    \label{fig:neurcomm_brkd}
\end{figure}
\begin{figure}[h]
    \centering
    \includegraphics[width=7.5cm]{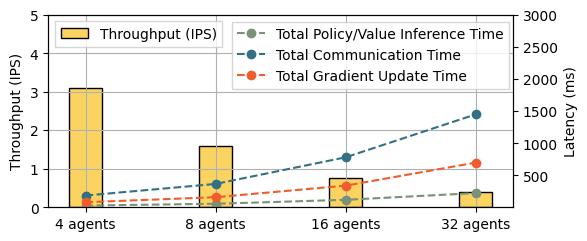}
    \caption{NeurComm Training Throughput Scalability varying the number of agents}
    \label{fig:neurcomm_scale}
\end{figure}

\begin{table*}
\small
\centering
\caption{Comparisons of target MARL algorithms. }
\label{tab:globalcomparison}
\begin{threeparttable}
\begin{tabular}{|c|c|c|c|}
\hline
\textbf{Algorithm} & \textbf{MADDPG} & \textbf{ToM2C} & \textbf{NeurComm} \\ \hline
\textit{Training Scheme} & Centralized & Centralized & Decentralized \\ \hline
\textit{Communication Method} & \begin{tabular}[c]{@{}c@{}}Pre-Defined,\\ All-to-all\end{tabular} & Learnt & \begin{tabular}[c]{@{}c@{}}Pre-Defined, \\ Graph-based\end{tabular} \\ \hline
\textit{$T_{Comm} \%$: Execution} & 5.89\% & 25.8\% & 72.2\% \\ \hline
\textit{$T_{Comm} \%$: Training} & 21.1\% & 47.0\% & 42.4\% \\ \hline
\textit{Parallel Rollout Support} & Yes & Yes & No \\ \hline
\textit{Parallel Training Support} & Yes & No & No \\ \hline
\textit{SG/MU Overlapping} & Yes & Yes & No \\ \hline
\textit{Cumulative Rewards} & \begin{tabular}[c]{@{}c@{}}CN\tnote{1} : -2.75+-0.61\\ ATSC\tnote{1} : Not Supported\end{tabular} & \begin{tabular}[c]{@{}c@{}}CN: -0.79+-0.39\\ ATSC: Not Supported\end{tabular} & \begin{tabular}[c]{@{}c@{}}CN: Not Supported\\  ATSC: -136.1\end{tabular} \\ \hline
\end{tabular}
\begin{tablenotes}
\item[1] Test Benchmarks: CN denotes Cooperative Navigation, ATSC denotes Adaptive Traffic Signal Control. \\
\item[2] $T_{Comm} \%$ stands for Communication Time Percentage. \\
\end{tablenotes}
\end{threeparttable}
\end{table*}
NeurComm-enabled MARL follows the Decentralized Training with Pre-Defined Communication paradigm aimed at targetting networked Multi-agent Reinforcement Learning (NMARL) scenarios, including traffic light and wireless networks systems \cite{chu2020multi}. This is formulated using a spatiotemporal Markov Decision Process, where each agent's actions are learned based on the state and policy of neighboring agents. The main contributions of this paper is a differentiable neural communication protocol called NeurComm. NeurComm introduces a message primitive called an agent's ``belief," which is propagated to the neighboring agents and optimizes their performance iteratively. The additional communication overhead lies in the belief propagation function. Each decentralized agent uses the belief, local states, and action probabilities of neighboring agents in order to make a decision. 

In this Decentralized NMARL using NeurComm communication implementation, A2C agents are used, which follow an on-policy actor-critic method \cite{mnih2016asynchronous}. Figure \ref{fig:neurcomm_brkd} shows the Training in Simulation breakdown of NeurComm during one iteration of Sample Generation and Model Update. During Sample Generation, messages are received and sent while control is performed. During Model Update, each agent's belief is updated along with the gradients of the actor, critic, and neural communication network. Overall, we observe that communication takes up a large percentage of the total execution time during an iteration of Sample Generation and Model Update.

NeurComm lacks a parallel setup and instead utilizes a single thread for both the Sample Generation and Model Update phases. From Figure \ref{fig:neurcomm_scale}, we observe a $2\times$ increase in all latency measurements when the number of agents doubles. When increasing the number of agents, various data structures, including belief, agent states, number of policies, etc., also increase in size proportionally. $IPS$ decreases proportionally with the number of agents, which is expected given the single-threaded nature of this implementation and the increasing communication and state size. While the addition of the belief primitive leads to optimized control performance, each agent needs to compute its own belief using the belief of other neighboring agents, leading to a large amount of communication and computation overhead.

\subsection{Comparison of MARL Algorithms}
Based on the analyses in Sections \ref{sec:maddpg}, \ref{sec:tom2c} and \ref{sec:neurcomm}, we summarize some tradeoffs between these different categories of algorithms and their parallel implementations in Table \ref{tab:globalcomparison}.

\textbf{Comparison in terms of Communication Methods:}
In CTDE (columns MADDPG and ToM2C of Table \ref{tab:globalcomparison}, row ``\textit{Cumulative Rewards}"), it has been verified that learning to communicate through predicting other agents' future rollouts outperforms simple all-to-all concatenation of current experiences. However, learnt communication also leads to higher communication cost in both decentralized execution (Sample generation) and centralized training (Model Update), as shown in 
% Figure \ref{fig:3d}, 
 Table \ref{tab:globalcomparison}, rows ``\textit{Communication Time Percentage},"
where the time spent on generating the message and data transfer for communication in ToM2C accounts for 3 times larger percentage than that in MADDPG. This calls for fine-grained acceleration of training to alleviate the training bottleneck in MARL frameworks supporting CTDE with learnt communication.

\textbf{Comparison in terms of Training Schemes:}
Note that Centralized training vs. Decentralized training implies different assumptions in terms of the accessibility of global information on the joint policy. Decentralized training can better support applications requiring autonomous acting using local information (e.g., traffic network, stock market), but each agent needs to cope with a non-stationary environment due to the instability of other changing and adapting agents. In
contrast, centralized training relies on a controlling authority of all agents' policies and faces a stationary environment. However, the large (exponentially increasing) joint policy space could be too difficult to search. We observe that in order to learn a stable policy, decentralized training relies more on communication due to the lack of global information. For instance, in NeurComm (Table \ref{tab:globalcomparison}, row ``\textit{Communication Time Percentage: Agent Execution}"), communication accounts for as much as 70\% of the total training time. The additional overhead involves DNN layer computation for the encoding and extraction of an agent's belief. This additional communication overhead justifies the need for fine-grain acceleration of the various added kernels.
\section{\uppercase{Discussion \& Conclusion}}
%\textcolor{blue}{$\sim$0.5 page}

In this work, we provided an extensive analysis of various MARL algorithms based on a taxonomy. This is also the first work in the field of MARL that characterizes key algorithms from a parallelization and acceleration perspective. We proposed the need for latency-bounded throughput to be considered a key optimization metric in future literature. Based on our observation, the need for communication brings a non-trivial overhead that needs fine-grained optimization and acceleration depending on the category of the algorithm described in our taxonomy. There is a plethora of future work that can be conducted on MARL in terms of acceleration: 
\begin{itemize}
    \item Specialized accelerator design for reducing communication overheads: Specialized acceleration platforms such as Field Programmable Gate Arrays (FPGA) offer pipeline parallelism along with large distributed on-chip memory that features single-cycle data access. To take full advantage of low-latency memory, specialized data layout and partition for the communicated message pool need to be exploited.
    \item Fine-grained Task Mapping using heterogeneous platforms: We have seen the success of bringing single-agent RL algorithm to heterogeneous platforms composed of CPU, GPU and FPGA \cite{meng2021ppoaccel,zhang2023framework} and plan to extend this to MARL.
\end{itemize}

\bibliographystyle{apalike}
{\small
\bibliography{ref}}

% \section*{\uppercase{Appendix}}

% If any, the appendix should appear directly after the
% references without numbering, and not on a new page. To do so please use the following command:
% \textit{$\backslash$section*\{APPENDIX\}}

\end{document}